\documentclass[10pt,twocolumn,letterpaper]{article}

\usepackage[pagenumbers]{cvpr} 

\usepackage{multirow}
\usepackage{colortbl}
\usepackage{xcolor}
\usepackage{subcaption}
\usepackage{balance}

\usepackage{xspace}
\makeatletter\renewcommand\paragraph{\@startsection{paragraph}{4}{\z@}
  {.15em \@plus1ex \@minus.2ex}{-.5em}{\normalfont\normalsize\bfseries}}\makeatother

\newcommand{\app}{\raise.17ex\hbox{$\scriptstyle\sim$}}
\def\eg{\emph{e.g}\onedot} 
\def\ie{\emph{i.e}\onedot}

\makeatother

\newlength\savewidth\newcommand\shline{\noalign{\global\savewidth\arrayrulewidth
  \global\arrayrulewidth 1pt}\hline\noalign{\global\arrayrulewidth\savewidth}}

\definecolor{lightgray}{rgb}{0.7421875,0.7421875,0.7421875}
\definecolor{deepgray}{gray}{.9}

\definecolor{cvprblue}{rgb}{0.21,0.49,0.74}
\usepackage[pagebackref,breaklinks,colorlinks,allcolors=cvprblue]{hyperref}

\usepackage{xspace}
\makeatletter\renewcommand\paragraph{\@startsection{paragraph}{4}{\z@}
  {.15em \@plus1ex \@minus.2ex}{-.5em}{\normalfont\normalsize\bfseries}}\makeatother

\def\paperID{14568} 
\def\confName{CVPR}
\def\confYear{2025}

\title{Double Visual Defense: Adversarial Pre-training and Instruction Tuning for Improving Vision-Language Model Robustness}

\author{
Zeyu Wang\textsuperscript{1,2*} \quad
Cihang Xie\textsuperscript{1} \quad
Brian Bartoldson\textsuperscript{2} \quad
Bhavya Kailkhura\textsuperscript{2} \quad \vspace{.5em}
\\
\textsuperscript{1}UC Santa Cruz  \quad\quad\quad
\textsuperscript{2}Lawrence Livermore National Laboratory
}

\begin{document}
\maketitle

\def\thefootnote{*}\footnotetext{Work done during an internship at LLNL.}

\begin{abstract}
This paper investigates the robustness of vision-language models against adversarial visual perturbations and introduces a novel ``double visual defense" to enhance this robustness. Unlike previous approaches that resort to lightweight adversarial fine-tuning of a pre-trained CLIP model, we perform large-scale adversarial vision-language pre-training from scratch using web-scale data. We then strengthen the defense by incorporating adversarial visual instruction tuning. The resulting models from each stage, $\Delta$CLIP and $\Delta^2$LLaVA, show substantially enhanced zero-shot robustness and set a new state-of-the-art in adversarial defense for vision-language models. For example, the adversarial robustness of $\Delta$CLIP surpasses that of the previous best models on ImageNet-1k by \app20\%.
Similarly, compared to prior art, $\Delta^2$LLaVA brings a \app30\% robustness improvement to image captioning task and a \app20\% robustness improvement to visual question answering task. Furthermore, our models exhibit stronger zero-shot recognition capability, fewer hallucinations, and superior reasoning performance compared to baselines. 
Our project page is \href{https://doublevisualdefense.github.io/}{https://doublevisualdefense.github.io/}.
\end{abstract}    
\section{Introduction}
\label{sec:intro}

\begin{figure*}[!t]
    \centering
    \includegraphics[scale=0.56]{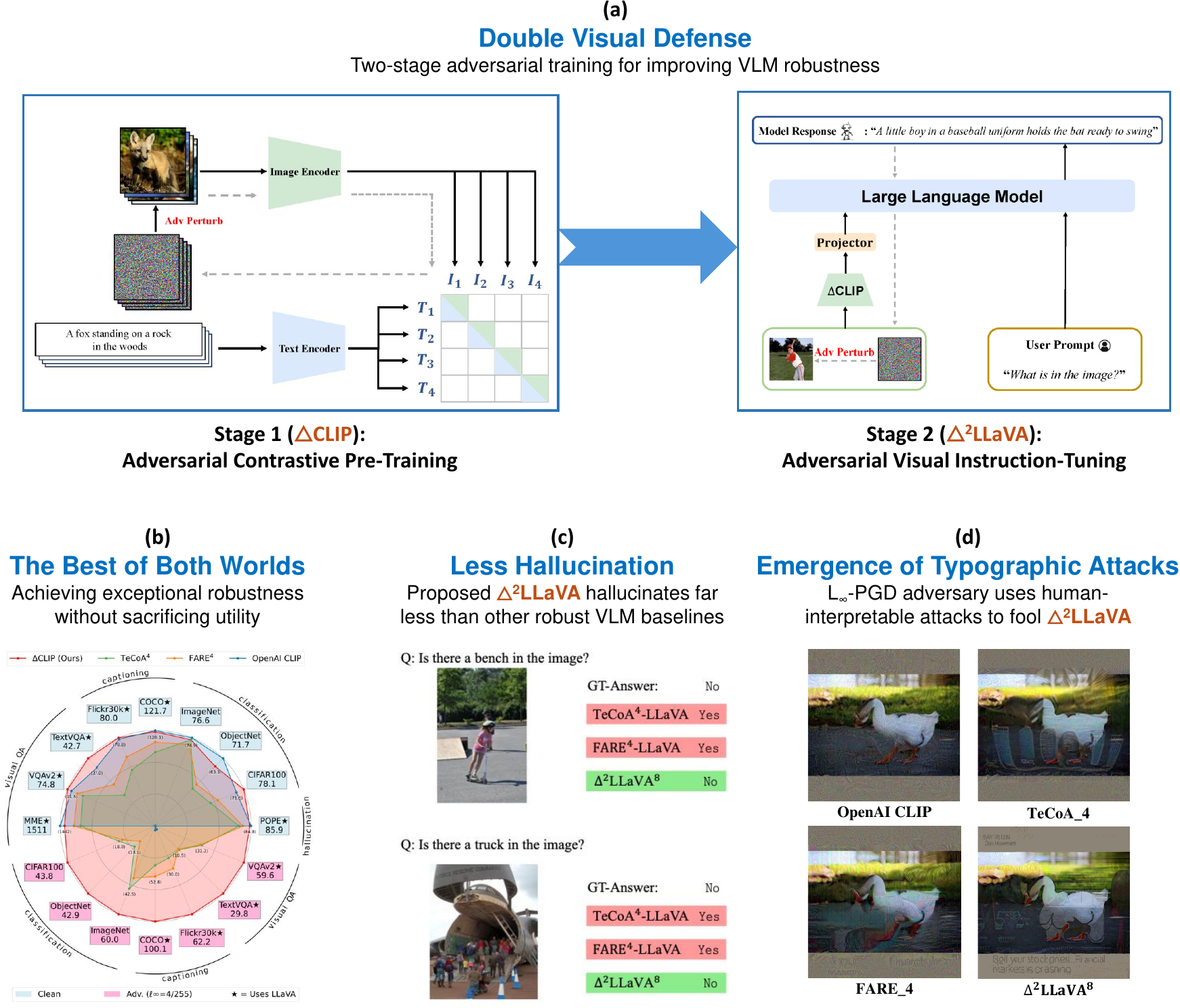}
    \caption{(a) Our Double Visual Defense framework, which involves an adversarial contrastive pre-training stage and an adversarial visual instruction tuning stage.  (b) Comparison of clean performance and robustness of our $\Delta$CLIP model with previous robust and non-robust CLIP models on 4 different tasks, including zero-shot recognition, image captioning, visual question answering, and hallucination. It can be seen that our $\Delta$CLIP attains drastically better robustness while maintaining clean performance close to that of the non-robust OpenAI CLIP counterpart. Note that our $\Delta^2$LLaVA shows further improved robustness upon $\Delta$CLIP on downstream VLM tasks (check section \ref{sec:adversarial_visual_instruction_tuning} and \ref{sec:experiments}). (c) $\Delta^2$LLaVA shows less degree of hallucination compared to LLaVA that are based on previous robust CLIP models like TeCoA \cite{mao2023understanding} or FARE \cite{schlarmann2024robust}. (d) We observe an intriguing phenomenon that typographical attack naturally emerge from naive $\ell_{\infty}$-adversarial attacks when applied to our adversarially trained $\Delta^2$LLaVA models. Best viewed when zoomed in.}
    \label{fig:teaser}
    \vspace{-.1em}
\end{figure*}

Vision-Language Models (VLMs) have become a crucial tool across domains, powering applications that bridge visual understanding and language comprehension \citep{liu2024visual,zhu2023minigpt,liu2024improved}.
A foundational innovation in this area is the CLIP model \cite{radford2021learning}, which connects visual and textual information within a unified embedding space by contrastive learning. Due to its excellent zero-shot recognition and generalization capability, CLIP has been widely used to empower the development of VLMs in various areas, including MiniGPT-4 \cite{zhu2023minigpt}, InstructBLIP \cite{dai2023instructblip} and LLaVA \cite{liu2024visual}. Notably, by integrating the CLIP visual encoder with a language decoder, these models enable open-set visual question answering, and more broadly speaking, a general-purpose instruction-following visual agent.

Despite these rapid and groundbreaking developments, these VLMs' susceptibility to visual adversarial attacks poses a persistent challenge. Adversarial perturbations, which are subtle and often imperceptible changes to input images, can drastically alter the output of VLMs, causing them to misinterpret or misclassify content \cite{goodfellow2014explaining,carlini2017towards}. In particular, a line of recent works has shown that VLMs are vulnerable to adversarial attacks \cite{zou2023universal,liu2024autodan,qi2023visual,carlini2024aligned}. In scenarios where VLMs might provide public information or guide user interactions, adversarial attacks could lead to the propagation of misinformation, defraud unsuspecting users, or compromise the integrity of automated decision-making systems \citep{qi2024visual,bailey2023image,wu2024safety}.

Efforts to improve the adversarial robustness of neural networks have introduced a range of adversarial training approaches \cite{Madry2018,shafahi2019adversarial,pang2021bag,zhang2020attacks,wang2024revisiting}, a process that involves generating adversarial examples on-the-fly for training. In this paper, we focus on helping VLMs defend against attacks on their visual channels. The most relevant works in this literature are TeCoA \cite{mao2023understanding} and FARE \cite{schlarmann2024robust}. Because CLIP training on Internet data is already computationally expensive and adversarial training adds often 5x to 10x more compute, both of them resort to a lightweight training stage that adapts a pre-trained CLIP vision encoder to make it resilient to adversarial attacks.  However, our experiments reveal that such quick fine-tuning on ``small" datasets (\eg ImageNet) might be prone to overfitting, hindering the zero-shot recognition and generalization ability of the original model.

Thus, in this paper, we investigate the following question: By switching from lightweight, post-hoc adversarial training approaches, to an approach that adversarially trains the VLM at all phases (CLIP pre-training and LLaVA instruction tuning), \textit{can we further improve adversarial robustness while preserving broad usefulness across uncorrupted inputs?} To investigate this question, we start by incorporating adversarial training into CLIP learning on web-scale data. Here, adversarial perturbations to visual inputs cause pairs of unrelated images and captions to match -- creating an adversarial version of the original CLIP's contrastive pre-training objective -- and our resulting CLIP model $\Delta$CLIP learns to defend against such attacks while achieving a well-aligned image-text embedding space.  Notably, we find that a LLaVA with a $\Delta$CLIP backbone has higher robustness than a LLaVA that uses the adversarially finetuned (and prior robustness SoTA) FARE model \cite{schlarmann2024robust}; however, we go further and add a second layer of defense by integrating adversarial autoregressive language modeling into the visual instruction tuning stage. By incorporating this second defense, which involves training on images perturbed to produce next token mispredictions, we further improve LLaVA model robustness to adversarial attacks. The combined approach is a Double Defense ($\Delta^2$) to attacks in the visual domain, and we accordingly name the resulting LLaVA model $\Delta^2$LLaVA.

Our contributions are summarized as follows:
\begin{enumerate}
    \item By switching from short-term post-hoc adversarial fine-tuning on ImageNet to our Double Visual Defense approach during both web-scale CLIP pre-training and visual instruction tuning, our models achieve superior robustness at little-to-no cost in clean data performance.
    \item To the best of our knowledge, we are the first to propose adversarial visual instruction tuning, and we find it benefits robustness, especially under strong attacks (see Section \ref{sec:llava_robustness_evaluation}).
    \item We test our models via a comprehensive evaluation on over 20 datasets and 4 evaluation setups, which offer a holistic understanding of the VLMs we train. Across all datasets, our ``$\Delta$'' series of models is either competitive with or far beyond prior works (see Figure \ref{fig:teaser} for illustration). For example, $\Delta$CLIP achieves an \app 70\% absolute robustness improvement (\app 700\% relative improvement) on Stanford Cars \cite{krause2013collecting} compared to other Robust CLIP models like TeCoA \cite{mao2023understanding} and FARE \cite{schlarmann2024robust}. Also, our $\Delta^2$LLaVA hallucinates far less and is much more robust compared to TeCoA-based and FARE-based LLaVAs.
    \item In sum, our VLMs are the first to reach non-robust-VLM helpfulness levels on clean data while being robust on adversarially attacked data. We believe our models can serve as drop-in replacements for vanilla CLIP and LLaVA in many cases, and we will release our code and model weights to benefit future VLM safety works.
\end{enumerate}

\section{Related Work}

\paragraph{Vision-Language Models} Since the arrival of Large Language Models (LLMs), a major research goal has been augmenting them with a visual skillset that complements their textual understanding and reasoning capabilities.
A seminal model in this area is CLIP \cite{radford2021learning}, one of the first works to connect vision and language learning by training on web-scale image-text pair data. Since CLIP's introduction, a number of followup works have sought to improve CLIP learning from model, data, learning strategy, and other perspectives \cite{yu2022coca,li2023scaling,li2024inverse,sun2023eva,li2024if}. Moreover, the superb zero-shot recognition and generalizability of CLIP has been pivotal in driving the development of next-generation VLMs. Among them, MiniGPT-4 \cite{zhu2023minigpt}, InstructBLIP \cite{dai2023instructblip} and LLaVA \cite{liu2024visual} are key illustrations of how CLIP can be used to equip LLMs with visual abilities. Specifically, by transforming the visual tokens from a pre-trained CLIP encoder into tokens in the LLM text embedding space, image and text tokens can be treated equally in an autoregressive modeling approach, resulting in models with both open-set visual recognition and language instruction-following and reasoning capabilities. In this work, we focus on improving the adversarial robustness of LLaVA and CLIP -- a widely adopted VLM and the backbone of its visual abilities, respectively.

\paragraph{Classical Adversarial Threats and Defenses} First discovered in \citet{szegedy2013intriguing}, adversarial examples cause neural networks to misbehave by adding small perturbations to the clean input, where the perturbations are found using the gradient of the loss with respect to the input. Subsequent work in this area has led to a series of stronger attacks  \cite{moosavi2016deepfool,carlini2017towards,Madry2018,eykholt2018robust,athalye2018obfuscated}. Adversarial training \cite{goodfellow2014explaining,Madry2018} has emerged as the key approach to defending against such attacks. It and its improved versions \cite{xie2019feature,shafahi2019adversarial,Wong2020Fast,zhang2021geometryaware,ijcai2019p595,wang2024revisiting,bartoldson2024adversarial} involve training on adversarial inputs generated on-the-fly during training. In this work, beyond traditional adversarial training on closed-set image classification tasks, we study how open-set VLM learning benefits from adversarial training.

\paragraph{VLM Adversarial Threats and Defenses} While adding visual reasoning abilities to LLMs to obtain VLMs has greatly advanced the scope of tasks and applications that large-scale models can address, it has also opened up a new security vulnerability: now, malicious attackers can initiate attacks from both vision and language channels \cite{schlarmann2023on,zou2023universal,carlini2024aligned,bailey2023image,qi2023visual,gu2024agent}. The attacks most relevant to our paper are those that make use of gradient information to craft malicious visual inputs that induce harmful or objectionable output \cite{bailey2023image,qi2023visual,schaeffer2024universal}. By definition, these adversarial attacks become more difficult when VLM robustness is improved. Accordingly, approaches to bolstering the adversarial defenses and thus safety/helpfulness of VLMs are of critical importance: TeCoA proposes text-based supervised adversarial fine-tuning, and FARE proposes feature-based unsupervised adversarial fine-tuning -- each method relies on a pre-trained CLIP model and ImageNet data \cite{mao2023understanding,schlarmann2024robust}. In this work, we avoid such lightweight and post-hoc adversarial adapting approaches, and we instead aim to study the effect of conducting adversarial learning at all phases of VLM training. We find that the result is drastically improved robustness and much better preservation of clean (not attacked) data performances.
\section{Methodology}
In this section, we introduce our Double Visual Defense framework, which integrates adversarial training into both CLIP pre-training and LLaVA instruction tuning to improve VLM robustness. In section \ref{sec:adversarial_training}, we first give a brief overview of adversarial training. In section \ref{sec:adversarial_clip_training}, we explain how we transform CLIP pretraining via an adversarial contrastive image-text matching objective.  In section \ref{sec:adversarial_visual_instruction_tuning}, we present our adversarial visual instruction tuning approach that builds on traditional LLaVA training. The resulting $\Delta$-series of models -- $\Delta$CLIP, $\Delta$LLaVA, and $\Delta^2$LLaVA -- exhibit state-of-the-art robustness while maintaining broad usefulness and helpfulness.

\subsection{Adversarial Training}
\label{sec:adversarial_training}
Adversarial examples are inputs designed to sabotage the usual decision making process of machine learning models. They are usually generated by adding small perturbations to regular data, like images. While these perturbations are typically subtle and not harmful to a human's ability to correctly recognize the original data, they nonetheless make models unreliable, causing them to make mispredictions, disregard their safety guardrails, etc.

Adversarial training is one of the most widely used defenses against adversarial examples. The core idea is to expose the model to adversarial examples during training to make the model less likely to be fooled by small perturbations, however well-crafted they are. Formally, given a network $\mathbf{f}_{\theta}$ with parameters $\theta$, adversarial training aims to optimize the following objective:

\begin{equation}
\min _\theta \max _{\|\delta\|_p \leq \epsilon_p} \mathcal{L}\left(\mathbf{f}_{\theta}(\mathbf{x}+\delta), \mathbf{y}\right).
\label{eq:adv train}
\end{equation}

Here we use $\mathbf{x}$ to denote an input image, $\delta$ to denote the additive adversarial perturbation, $\mathbf{y}$ to denote the label, and $\mathcal{L}$ to denote the loss function.

\subsection{Adversarial Contrastive Language-Image Pretraining}
\label{sec:adversarial_clip_training}
The CLIP model learns a well-aligned image-text joint embedding space by training an image encoder and a text encoder to predict the correct image-text associations. By learning on web-scale data that is rich with natural language supervision, it transcends pre-defined categories and generalizes well across different tasks and domains in an ``out of the box" fashion, making it effective for open-vocabulary visual recognition. Specifically, the contrastive loss in CLIP training can be formulated as

\begin{equation}
\begin{split}
& \mathcal{L}_{con} \left(\mathbf{x},  \mathbf{y} \right) =  \\
& - \mathbb{E}_{(\mathbf{x}_{i}, \mathbf{y}_{j})} \left[\mathbf{m}_{ij} \log \frac{\exp\left( \cos\left( \mathbf{f}_{\theta^{I}}(\mathbf{x}_{i}), \mathbf{f}_{\theta^{T}}(\mathbf{y}_{j}) \right)/ \tau\right)}{\sum_k \exp\left( \cos\left( \mathbf{f}_{\theta^{I}}(\mathbf{x}_{i}), \mathbf{f}_{\theta^{T}}(\mathbf{y}_{k}) \right)/ \tau\right)} \right. \\
& \left. + \mathbf{m}_{ij} \log \frac{\exp\left( \cos\left( \mathbf{f}_{\theta^{I}}(\mathbf{x}_{i}), \mathbf{f}_{\theta^{T}}(\mathbf{y}_{j}) \right)/ \tau\right)}{\sum_k \exp\left( \cos\left( \mathbf{f}_{\theta^{I}}(\mathbf{x}_{k}), \mathbf{f}_{\theta^{T}}(\mathbf{y}_{j}) \right)/ \tau\right)}  \right].
\end{split}
\label{eq:contrastive loss}
\end{equation}

In Equation \ref{eq:contrastive loss}, $\mathbf{x}$ is a batch of input images; $\mathbf{y}$ is a batch of input texts; $\mathbf{f}_{\theta^{I}}(\mathbf{x_i})$ is the feature vector of image $\mathbf{x_i}$ extracted by the vision encoder $\mathbf{f}_{\theta^{I}}$, $\mathbf{f}_{\theta^{T}}(\mathbf{y_j})$ is the feature vector of text $\mathbf{y_j}$ extracted by the text encoder $\mathbf{f}_{\theta^{T}}$; $\mathbf{m}_{ij}$ indicates whether an image-text pair is a match or not, $\mathbf{m}_{ij} = 1$ if and only if $i = j$ and is $0$ otherwise; $\tau$ is a learnable temperature parameter; and $\cos$ denotes the cosine similarity function.

Despite CLIP's great performance on open-set visual tasks, CLIP-based VLMs are highly vulnerable to adversarial attacks \cite{schlarmann2023on,qi2023visual,carlini2024aligned}, casting doubt on the ability to safely and responsibly deploy such models. To our knowledge, the only two previous works that try to robustify CLIP models resort to short-term post-hoc adversarial tuning on ImageNet \cite{mao2023understanding,schlarmann2024robust}. However, our experiments reveal that such a lightweight approach causes large performance drops in CLIP models on uncorrupted inputs, hindering such models' overall usefulness and helpfulness (see section \ref{sec:experiments}). 

In this paper, we instead conduct adversarial training from the start of CLIP's pretraining process to produce $\Delta$CLIP, which maintains CLIP's excellent zero-shot generalizability but significantly boosts its robustness. Notably, these robustness benefits are also visible in downstream CLIP-based VLMs (like LLaVA) that use our $\Delta$CLIP model as a visual backbone. Our approach is simple: $\Delta$CLIP is trained to predict the right image-text pairings given adversarial images that are optimized to fool the model into predicting incorrect image-text pairings. Formally, this process can be described as

\begin{equation}
\min _{\theta^{I}} \max _{\|\delta\|_p \leq \epsilon_p} \mathcal{L}_{con}\left(\mathbf{x}+\delta, \mathbf{y}\right).
\label{eq:adv_train_contrastive}
\end{equation}

\begingroup
\renewcommand{\arraystretch}{1.4} 
\begin{table*}[htbp]
     \centering
      \resizebox{.9\linewidth}{!}{
     \begin{tabular}{ccc|cccccc|cccc}
     \toprule
     &  & &\multicolumn{6}{c|}{\textbf{zero-shot classification}} &\multicolumn{4}{c}{\textbf{zero-shot retrieval}} \\
     &  & &\multirow{2}{*}{IN-1k} &\multirow{2}{*}{IN-V2} &\multirow{2}{*}{IN-A} &\multirow{2}{*}{IN-R} &\multirow{2}{*}{ObjectNet} &\multirow{2}{*}{IN-Sketch} &\multicolumn{2}{c}{COCO} &\multicolumn{2}{c}{Flickr30k} \\
     \textbf{Eval}     &\textbf{Model} &\textbf{Training Data} &     &  &  &  &  &  &image  &text  &image  &text \\ \shline
     \parbox[t]{4mm}{\multirow{8}{*}{\rotatebox[origin=c]{90}{\large{\textit{clean}}}}}
     &{\color{lightgray}OpenAI-L/14}     &{\color{lightgray}WIT-400M}       &{\color{lightgray}75.5} &{\color{lightgray}69.8}  &{\color{lightgray}70.8}  &{\color{lightgray}87.8}  &{\color{lightgray}68.9}  &{\color{lightgray}59.6}  &{\color{lightgray}36.5}  &{\color{lightgray}56.4}  &{\color{lightgray}65.3}  &{\color{lightgray}85.1} \\
     &{\color{lightgray}OpenAI-L/14-336} &{\color{lightgray}WIT-400M}       &{\color{lightgray}76.6} &{\color{lightgray}70.9}  &{\color{lightgray}77.5}  &{\color{lightgray}89.1}  &{\color{lightgray}71.7}  &{\color{lightgray}61.0}  &{\color{lightgray}37.1}  &{\color{lightgray}58.0}  &{\color{lightgray}67.3}  &{\color{lightgray}87.4} \\
     &{\color{lightgray}OpenCLIP-L/14}   &{\color{lightgray}LAION-400M}     &{\color{lightgray}72.8} &{\color{lightgray}65.4}  &{\color{lightgray}46.5}  &{\color{lightgray}84.9}  &{\color{lightgray}59.9}  &{\color{lightgray}59.6}  &{\color{lightgray}43.0}  &{\color{lightgray}59.7}  &{\color{lightgray}70.3}  &{\color{lightgray}87.6}  \\
     &TeCoA$^2$-L/14  &WIT-400M+ImageNet-1K        &\textbf{80.1} &\textbf{70.5}  &32.5  &80.1  &47.6  &58.4  &32.9  &40.3  &60.3  &69.8 \\
     &FARE$^2$-L/14   &WIT-400M+ImageNet-1K        &74.5 &67.3  &40.6  &85.5  &53.4  &59.7  &38.6  &53.6  &68.5  &84.1 \\
     &TeCoA$^4$-L/14  &WIT-400M+ImageNet-1K        &74.9 &64.1  &19.8  &74.4  &39.6  &54.2  &27.8  &32.9  &53.0  &58.5 \\
     &FARE$^4$-L/14   &WIT-400M+ImageNet-1K        &70.8 &62.2  &23.7  &80.2  &43.9  &56.7  &34.2  &45.9  &54.0  &77.6 \\
     &\cellcolor{deepgray} $\Delta$CLIP-H/14-336 &\cellcolor{deepgray} DataComp-1B     &\cellcolor{deepgray}74.8 &\cellcolor{deepgray}66.7  &\cellcolor{deepgray}\textbf{46.1}  &\cellcolor{deepgray}\textbf{91.3}  &\cellcolor{deepgray}\textbf{63.3}  &\cellcolor{deepgray}\textbf{68.3}  &\cellcolor{deepgray}\textbf{49.2}  &\cellcolor{deepgray}\textbf{68.4}  &\cellcolor{deepgray}\textbf{75.5}  &\cellcolor{deepgray}\textbf{90.7} \\ \hline
     \parbox[t]{4mm}{\multirow{8}{*}{\rotatebox[origin=c]{90}{\large $\ell_\infty = \frac{4}{255}$}}}
     &{\color{lightgray}OpenAI-L/14-336} &{\color{lightgray}WIT-400M}       &{\color{lightgray}0} &{\color{lightgray}0}  &{\color{lightgray}0}  &{\color{lightgray}0}  &{\color{lightgray}0}  &{\color{lightgray}0}  &{\color{lightgray}-}  &{\color{lightgray}-}  &{\color{lightgray}-}  &{\color{lightgray}-} \\
      &TeCoA$^2$-L/14  &WIT-400M+ImageNet-1K        &35.7 &22.7  &2.1  &36.7  &9.7  &32.6  &-  &-  &-  &- \\
     &FARE$^2$-L/14   &WIT-400M+ImageNet-1K        &17.4 &10.7  &1.2  &25.9  &4.7  &22.3  &-  &-  &-  &- \\
      &TeCoA$^4$-L/14  &WIT-400M+ImageNet-1K       &42.5 &30.6  &3.0  &41.9  &13.1  &34.3  &-  &-  &-  &- \\
     &FARE$^4$-L/14   &WIT-400M+ImageNet-1K        &35.4 &23.3  &2.6  &40.7  &9.7  &30.9  &-  &-  &-  &- \\
     &\cellcolor{deepgray} $\Delta$CLIP-H/14-336    &\cellcolor{deepgray} DataComp-1B     &\cellcolor{deepgray} \textbf{60.0} &\cellcolor{deepgray} \textbf{49.4}  &\cellcolor{deepgray} \textbf{21.6}  &\cellcolor{deepgray} \textbf{81.5}  &\cellcolor{deepgray} \textbf{42.9}  &\cellcolor{deepgray} \textbf{57.4}  &\cellcolor{deepgray} -  &\cellcolor{deepgray} -  &\cellcolor{deepgray} -  &\cellcolor{deepgray} - \\
     \bottomrule
     \end{tabular}}
     \captionof{table}{\textbf{Clean and adversarial zero-shot CLIP evaluation}. TeCoA and FARE are OpenAI CLIP models further finetuned on ImageNet-1K data. The clean OpenAI CLIP is completely non-robust despite its strong clean performances. The TeCoA and FARE models exhibit good robustness, but suffer from significant clean performance drops. By contrast, our $\Delta$CLIP shows both strong clean and adversarial performances. }
     \label{tab:zero_shot1}
     \vspace{-.4em}
\end{table*}
\endgroup

\subsection{Adversarial Visual Instruction Tuning}
\label{sec:adversarial_visual_instruction_tuning}

CLIP's ability to empower LLMs with open-set visual understanding has been demonstrated by various VLMs \cite{zhu2023minigpt, dai2023instructblip,liu2024visual}. However, the ability to corrupt and control these VLMs through adversarial attacks on their visual input \cite{bailey2023image,qi2023visual,schaeffer2024universal} makes improving their robustness crucial. 
Prior work suggested that use of a more robust CLIP model will make the downstream VLM more robust \cite{schlarmann2024robust}.  
However, as instruction fine-tuning itself can be harmful to the safety alignment of LLMs or VLMs \cite{qi2024finetuning,pantazopoulos2024learning}, we consider the possibility that adversarial training of the VLM may further improve robustness, even when the VLM already uses the visual encoder of a robust CLIP model like $\Delta$CLIP.

Indeed, beyond evaluating the performance of a LLaVA \cite{liu2024visual} that uses a $\Delta$CLIP visual encoder ($\Delta$LLaVA), we also perform adversarial LLaVA training to potentially achieve a second layer of defense. Specifically, we train both  $\Delta$LLaVA and  $\Delta^2$LLaVA -- the former relies only on the robustness of $\Delta$CLIP to defend against adversarial attacks, while the latter has the additional defense provided by our novel adversarial visual instruction tuning approach. 

Formally, given VLM parameters $\phi$, an image $\mathbf{x}$, and a string $\mathbf{y}$ that contains $L$ instruction and $L'$ target answer tokens, the baseline autoregressive loss used for LLaVA visual instruction tuning \cite{liu2024visual} can be expressed as

\begin{equation}
\mathcal{L}_{inst} \left(\mathbf{x},  \mathbf{y} \right) = -\sum_{t=L'}^{L+L'} \log p_{\phi}(\mathbf{y}_t | \mathbf{f}_{\theta^{I}}(\mathbf{x}), \mathbf{y}_{<t}).
\label{eq:visual_instruction_loss}
\end{equation}

As can be seen in Table \ref{tab:llava_robustness}, the robustness of a downstream VLM is greatly enhanced by use of visual features extracted from our $\Delta$CLIP model. However, we see further improvements when adding adversarial visual instruction tuning to grant the VLM a Double Defense against adversarial attacks. Concretely, adversarial visual instruction tuning adds adversarial noise within a perturbation radius to the input image to force the model to predict the wrong next token, and the model is trained to make the correct token predictions despite these perturbations. This adversarial autoregressive training process can be formulated as

\begin{equation}
\min _{\phi} \max _{\|\delta\|_p \leq \epsilon_p} \mathcal{L}_{inst}\left(\mathbf{x}+\delta, \mathbf{y}\right).
\label{eq:adv_visual_instruction_loss}
\end{equation}

Note that $\mathbf{f}_{\theta^{I}}(\mathbf{x})$ is added in the condition to highlight the fact that image is grounded for all answers.

We emphasize that, while prior works have attempted to defend against VLM adversarial examples by additive random noise or JPEG Compression \cite{bailey2023image}, our approach constitutes the first attempt to robustify VLMs via adversarial autoregressive training (to the best of our knowledge). It is also worth mentioning that we have tried adversarial visual instruction tuning on both vanilla CLIP-based and $\Delta$CLIP-based LLaVA models. The former attempt results in a completely crashed model, while the latter leads to stronger adversarial robustness, suggesting the importance of adversarial pre-training.
\section{Experiments}
\label{sec:experiments}

Following previous robust CLIP works \cite{mao2023understanding,schlarmann2024robust}, we evaluate the clean performance and adversarial robustness of the CLIP and LLaVA models produced by our approach. CLIP zero-shot performances are reported in Section \ref{sec:clip_zero_shot_recognition}. We then evaluate the clean and robust performances of LLaVA models on image captioning and visual question answering tasks in Section \ref{sec:llava_robustness_evaluation}. Next, in Section \ref{sec:target_attack_on_llava}, we evaluate how well different LLaVA models defend against targeted attacks that force the model to generate the exact output a malicious attacker desires. Finally, we probe the clean performances of these LLaVA models on visual reasoning and hallucination benchmarks in Section \ref{sec:llava_qa_performance} to see if they remain useful and helpful after being robustified.

\paragraph{Training Details} We train our $\Delta$CLIP model on the internet-crawled data DataComp-1B \cite{gadre2024datacomp}. We adopt the synthetic captions from Recap-DataComp-1B \cite{li2024if}, mixing them together with the original web captions at a 1:1 ratio for richer language supervision. The text model is pre-trained using clean data with the same schedule and kept frozen during adversarial training. We also incorporate the captioning loss from CoCa \cite{yu2022coca} in our adversarial pre-training framework as we observe in our early experiments that it is beneficial for both clean performances and robustness.

Following prior efficient CLIP training practices \cite{li2024inverse,wang2024revisiting}, we divide our CLIP training into three stages. In the first stage the model is trained with 112$\times$112 input image size and PGD-2 adversarial training. In the second stage the model is trained with 224$\times$224 input image size and PGD-3 adversarial training. Lastly, to match with the input image size used in LLaVA-1.5, we further train the CLIP model with 336$\times$336 input image size and PGD-4 adversarial training. In the first two stages, the attack radius $\epsilon=4/255$ is used. In the third stage, the attack radius $\epsilon=8/255$ is used. The model was trained on about 5.12B, 512M, and 128M samples during each stage, respectively. 

We adopt the LLaVA-1.5 training recipe \cite{liu2024improved} across the whole paper. Low-Rank Adaptation (LoRA) \cite{hu2022lora} is adopted when training $\Delta^2$LLaVA to lower cost. We use two attacks, PGD-3 under radius $4/255$ and PGD-5 under radius $8/255$ in adversarial visual instruction tuning, and name the resulting models $\Delta^2$LLaVA$^4$ and $\Delta^2$LLaVA$^8$, respectively. Note that in the original LLaVA-1.5 training recipe, the vision encoder remains frozen even in the fine-tuning stage, but we instead keep the learning rate of the vision encoder at $\frac{1}{20}$ the base learning rate in adversarial fine-tuning.

Our CLIP model is implemented based on JAX \cite{bradbury2018jax} and run on TPU v4 infrastructure. The $\Delta$CLIP-H/14-336 model took about a week to finish on a TPU v4-512 pod.
Our LLaVA model is implemented based on PyTorch \cite{bradbury2018jax} and run on NVIDIA A5000/A100 and AMD MI250X GPU infrastructure. The $\Delta^2$LLaVA$^8$ model was trained on 4 8XA5000 GPU machines for about 1.5 days. 

\begingroup
\renewcommand{\arraystretch}{1.5} 
\begin{table*}[htbp]
    \centering
    \resizebox{.75\linewidth}{!}{
        \begin{tabular}{ccc|ccccccccc}
        \toprule
        \textbf{Eval}
        &\textbf{Model}
        &\textbf{Training Data}
        & \rotatebox[origin=l]{90}{Caltech101}
        & \rotatebox[origin=l]{90}{Cars}
        & \rotatebox[origin=l]{90}{Cifar10}
        & \rotatebox[origin=l]{90}{Cifar100}
        & \rotatebox{90}{Dtd}
        & \rotatebox{90}{Eurosat}
        & \rotatebox{90}{FGVC}
        & \rotatebox{90}{Flowers}
        & \rotatebox{90}{Pets} \\ \shline
        \parbox[t]{4mm}{\multirow{8}{*}{\rotatebox[origin=c]{90}{\large{\textit{clean}}}}}
        &{\color{lightgray}OpenAI-L/14}     &{\color{lightgray}WIT-400M}       &{\color{lightgray}83.3} &{\color{lightgray}77.9}  &{\color{lightgray}95.6}  &{\color{lightgray}75.8}  &{\color{lightgray}55.3}  &{\color{lightgray}62.6}  &{\color{lightgray}31.6}  &{\color{lightgray}79.2}  &{\color{lightgray}93.2} \\
        &{\color{lightgray}OpenAI-L/14-336}     &{\color{lightgray}WIT-400M}       &{\color{lightgray}83.4} &{\color{lightgray}79.4}  &{\color{lightgray}94.9}  &{\color{lightgray}74.4}  &{\color{lightgray}55.7}  &{\color{lightgray}61.4}  &{\color{lightgray}33.3}  &{\color{lightgray}78.2}  &{\color{lightgray}93.6} \\
        &{\color{lightgray}OpenCLIP-L/14}     &{\color{lightgray}LAION-400M}       &{\color{lightgray}84.0} &{\color{lightgray}89.6}  &{\color{lightgray}94.7}  &{\color{lightgray}77.4}  &{\color{lightgray}60.5}  &{\color{lightgray}62.3}  &{\color{lightgray}25.0}  &{\color{lightgray}75.6}  &{\color{lightgray}91.9} \\
        &TeCoA$^2$-L/14  &WIT-400M+ImageNet-1K        &80.7 &50.2  &86.9  &59.4  &44.4  &26.0  &14.1  &51.8  &80.1  \\
        &FARE$^2$-L/14   &WIT-400M+ImageNet-1K        &84.7 &70.5  &89.0  &68.2  &49.8  &25.3  &26.7  &70.6  &91.7  \\
        &TeCoA$^4$-L/14  &WIT-400M+ImageNet-1K       &78.4 &37.8  &78.4  &48.8  &38.0  &22.5  &11.8  &38.4  &76.1 \\
        &FARE$^4$-L/14   &WIT-400M+ImageNet-1K        &84.7 &63.8  &76.3  &55.2  &43.8  &18.2  &22.0  &58.0  &87.1  \\
        &\cellcolor{deepgray} $\Delta$CLIP-H/14-336    &\cellcolor{deepgray}DataComp-1B     &\cellcolor{deepgray}\textbf{85.1} &\cellcolor{deepgray}\textbf{91.7}  &\cellcolor{deepgray}\textbf{95.1}  &\cellcolor{deepgray}\textbf{78.1}  &\cellcolor{deepgray}\textbf{60.0}  &\cellcolor{deepgray}\textbf{37.8}  &\cellcolor{deepgray}\textbf{40.3} &\cellcolor{deepgray}\textbf{77.0}  &\cellcolor{deepgray}\textbf{92.1}  \\ \hline
        \parbox[t]{4mm}{\multirow{8}{*}{\rotatebox[origin=c]{90}{\large $\ell_\infty = \frac{4}{255}$}}}
        &{\color{lightgray}OpenAI-L/14-336}     &{\color{lightgray}WIT-400M}       &{\color{lightgray}0} &{\color{lightgray}0}  &{\color{lightgray}0}  &{\color{lightgray}0}  &{\color{lightgray}0}  &{\color{lightgray}0}  &{\color{lightgray}0}  &{\color{lightgray}0}  &{\color{lightgray}0} \\
        &TeCoA$^2$-L/14  &WIT-400M+ImageNet-1K        &57.1 &6.5  &19.9  &11.7  &14.6  &7.7  &1.1  &9.3  &50.5  \\
        &FARE$^2$-L/14   &WIT-400M+ImageNet-1K        &45.7 &5.0  &12.1  &7.8  &11.8  &0.3  &0.6  &7.0  &28.3  \\
        &TeCoA$^4$-L/14  &WIT-400M+ImageNet-1K       &61.0 &8.5  &29.7  &18.0  &16.8  &6.5  &2.0  &12.4  &55.2 \\
        &FARE$^4$-L/14   &WIT-400M+ImageNet-1K       &64.0 &12.7  &27.2  &16.3  &17.3  &\textbf{11.1}  &2.4  &12.2  &50.8  \\
        &\cellcolor{deepgray} $\Delta$CLIP-H/14-336    & \cellcolor{deepgray}DataComp-1B     &\cellcolor{deepgray}\textbf{80.4} &\cellcolor{deepgray}\textbf{88.0}  &\cellcolor{deepgray}\textbf{68.0}  &\cellcolor{deepgray}\textbf{43.8}  &\cellcolor{deepgray}\textbf{45.4}  &\cellcolor{deepgray}5.0  &\cellcolor{deepgray}\textbf{30.4} &\cellcolor{deepgray}\textbf{66.8}  &\cellcolor{deepgray}\textbf{78.6}  \\
        \bottomrule
        \end{tabular}
        }
      \caption{\textbf{More clean and adversarial zero-shot CLIP evaluation}. TeCoA and FAR are OpenAI CLIP models further finetuned on ImageNet-1K data. The clean OpenAI CLIP model is completely non-robust despite its strong clean performances. The TeCoA and FARE models suffer from significant performance drops on non-ImageNet-variant data. By contrast, our $\Delta$CLIP shows strong adversarial performances while maintaining the good generalizability of CLIP models. }
    \label{tab:zero_shot2}
\end{table*}
\endgroup

\subsection{CLIP Zero-Shot Recognition}
\label{sec:clip_zero_shot_recognition}

\paragraph{Evaluation Setup} Similar to \citet{mao2023understanding}, we compare the performance of our $\Delta$CLIP model against other CLIP models on a broad range of zero-shot benchmarks to reflect their relative generalization capabilities.
We follow the standard prompt engineering template in \verb|CLIP_benchmark|\footnote{https://github.com/LAION-AI/CLIP\_benchmark} to generate the text embedding for each class. When evaluating zero-shot adversarial robustness, we follow \citet{schlarmann2024robust} and opt for APGD-100 with cross entropy loss plus APGD-100 with DLR loss as in AutoAttack \cite{croce2020reliable}. The robustness is evaluated on 1000 random samples from each dataset and clean performance is evaluated on all samples in each dataset. Note that our random selection is different from that in \citet{schlarmann2024robust}, and thus the results reported in their paper are not directly comparable to ours.

\paragraph{Results} The results are shown in Table \ref{tab:zero_shot1} and Table \ref{tab:zero_shot2}. As can be seen, our $\Delta$CLIP model achieves on par or even better performance on clean data compared to the non-robust OpenAI CLIP and OpenCLIP models. Note that the OpenAI CLIP model was trained on a private dataset WIT-400M \cite{radford2021learning}, and it tends to produce favorable performance on certain datasets like ImageNet-A \cite{li2023scaling}. It also can be observed that while the TeCoA and FARE models seem to do fine on ImageNet, the lightweight adversarial tuning process results in significant performance decrease on other datasets. For example, on ImageNet-A, TeCoA$^4$'s and FARE$^4$'s adversarial adapting of the OpenAI CLIP model leads to \app 50\% and \app 45\% absolute performance drops on ImageNet-A, respectively. A similar accuracy decrease of \app 30\% happens on ObjectNet. The evaluation on non-ImageNet-variant datasets further corroborates the superiority of $\Delta$CLIP. For instance, the robustness of $\Delta$CLIP surpasses that of the second best model (FARE$^4$) by \app 75\% on the Stanford Cars dataset, boosting the accuracy almost $7\times$. To explain such phenomena, we hypothesize that the post-hoc adversarial fine-tuning approach leads to severe overfitting to ImageNet, due to its fine-tuning data's lack of diversity and richness. Contrastingly, $\Delta$CLIP was adversarially trained on diverse data and is the only high-performing robust CLIP model in this setting.

\begingroup
\renewcommand{\arraystretch}{1.2} 
\begin{table*}[htbp]
     \centering
     \resizebox{.6\linewidth}{!}{
     \begin{tabular}{ccc|cccc}
     \toprule
     \textbf{Eval}      &\textbf{Model} &\textbf{Vision Encoder} &COCO  &Flickr30k  &VQAv2  &TextVQA \\ \shline
     \parbox[t]{4mm}{\multirow{6}{*}{\rotatebox[origin=c]{90}{\large{\textit{clean}}}}}
     &\multirow{3}{*}{LLaVA}
     &{\color{lightgray}OpenAI-L/14}    &{\color{lightgray}121.7}  &{\color{lightgray}78.8}  &{\color{lightgray}71.5}  &{\color{lightgray}37.0} \\
     & &TeCoA$^4$-L/14  &94.0  &50.9  &61.8  &19.5 \\
     & &FARE$^4$-L/14   &106.8  &62.6  &66.4  &26.2 \\
     &\cellcolor{deepgray} $\Delta$LLaVA &\cellcolor{deepgray} $\Delta$CLIP-H/14-336 &\cellcolor{deepgray} \textbf{120.1}  &\cellcolor{deepgray} \textbf{80.0}  &\cellcolor{deepgray} \textbf{74.8}  &\cellcolor{deepgray} 42.7 \\
     &\cellcolor{deepgray}$\Delta^2$LLaVA$^4$ &\cellcolor{deepgray} $\Delta$CLIP-H/14-336 &\cellcolor{deepgray} 116.6  &\cellcolor{deepgray} 78.0  &\cellcolor{deepgray} 73.8  &\cellcolor{deepgray} \textbf{43.3} \\
     &\cellcolor{deepgray} $\Delta^2$LLaVA$^8$ &\cellcolor{deepgray} $\Delta$CLIP-H/14-336 &\cellcolor{deepgray} 108.6 &\cellcolor{deepgray} 66.0  &\cellcolor{deepgray} 69.5  &\cellcolor{deepgray} 38.2 \\ \hline
     \parbox[t]{4mm}{\multirow{6}{*}{\rotatebox[origin=c]{90}{\large $\ell_\infty = \frac{4}{255}$}}}
     &\multirow{3}{*}{LLaVA}
     &{\color{lightgray}OpenAI-L/14}    &{\color{lightgray}4.4}  &{\color{lightgray}2.3}  &{\color{lightgray}0.3}  &{\color{lightgray}0.0} \\
     & &TeCoA$^4$-L/14  &41.1  &22.5  &31.3  &10.5 \\
     & &FARE$^4$-L/14   &52.8  &30.0  &30.5  &10.2 \\
     &\cellcolor{deepgray} $\Delta$LLaVA &\cellcolor{deepgray} $\Delta$CLIP-H/14-336 &\cellcolor{deepgray} 100.1  &\cellcolor{deepgray} 62.2  &\cellcolor{deepgray} 59.6  &\cellcolor{deepgray} 29.8 \\
     &\cellcolor{deepgray} $\Delta^2$LLaVA$^4$ &\cellcolor{deepgray} $\Delta$CLIP-H/14-336 &\cellcolor{deepgray} \textbf{104.4}  &\cellcolor{deepgray} \textbf{63.3}  &\cellcolor{deepgray} \textbf{64.9}  &\cellcolor{deepgray} 31.1 \\
     &\cellcolor{deepgray} $\Delta^2$LLaVA$^8$ &\cellcolor{deepgray} $\Delta$CLIP-H/14-336 &\cellcolor{deepgray} 95.4  &\cellcolor{deepgray} 57.0  &\cellcolor{deepgray} 61.0  &\cellcolor{deepgray} \textbf{32.4} \\ \hline
     \parbox[t]{4mm}{\multirow{6}{*}{\rotatebox[origin=c]{90}{\large $\ell_\infty = \frac{8}{255}$}}}
     &\multirow{3}{*}{LLaVA}
     &{\color{lightgray}OpenAI-L/14}    &{\color{lightgray}2.6}  &{\color{lightgray}1.1}  &{\color{lightgray}0.0}  &{\color{lightgray}0.0} \\
     & &TeCoA$^4$-L/14  &25.0  &14.9  &22.4  &5.1 \\
     & &FARE$^4$-L/14   &30.4  &16.4  &20.7  &4.2 \\
     &\cellcolor{deepgray} $\Delta$LLaVA &\cellcolor{deepgray} $\Delta$CLIP-H/14-336 &\cellcolor{deepgray} 76.4  &\cellcolor{deepgray} 44.6  &\cellcolor{deepgray} 41.1  &\cellcolor{deepgray} 17.7 \\
     &\cellcolor{deepgray} $\Delta^2$LLaVA$^4$ &\cellcolor{deepgray} $\Delta$CLIP-H/14-336 &\cellcolor{deepgray} 79.8  &\cellcolor{deepgray} \textbf{47.2}  &\cellcolor{deepgray} 43.9  &\cellcolor{deepgray} 19.4 \\
     &\cellcolor{deepgray} $\Delta^2$LLaVA$^8$ &\cellcolor{deepgray} $\Delta$CLIP-H/14-336 &\cellcolor{deepgray} \textbf{81.3}  &\cellcolor{deepgray} 45.0  &\cellcolor{deepgray} \textbf{52.4}  &\cellcolor{deepgray} \textbf{22.5} \\
     \bottomrule
     \end{tabular}
     }
     \vspace{-.5em}
     \captionof{table}{\textbf{Evaluation of LLaVA robustness on image captioning and visual question answering tasks}. $\Delta$LLaVA that is trained with the vision encoder of $\Delta$CLIP surpasses TeCoA- and FARE-based LLaVA by a large margin in terms of robustness while maintaining clean performance close to the vanilla LLaVA model. And $\Delta^2$LLaVA further improves robustness upon $\Delta$LLaVA particularly with large radius attack.}
     \label{tab:llava_robustness}
     \vspace{-.5em}
\end{table*}
\endgroup

\subsection{LLaVA Untargeted Robustness Evaluation}
\label{sec:llava_robustness_evaluation}

\paragraph{Evaluation Setup} 
We follow \citet{schlarmann2024robust} and evaluate clean and robust performances on the COCO \cite{lin2014microsoft} and Flickr30k \cite{plummer2015flickr30k} datasets for the image captioning task, and on the VQAv2 \cite{goyal2017making} and TextVQA \cite{singh2019towards} datasets for the visual question answering (VQA) task. For all  tasks, 500 random samples are used for the adversarial evaluations, and all available samples are used for the clean evaluations. The CIDEr score \cite{vedantam2015cider} is used as the evaluation metric for image captioning and VQA accuracy \cite{antol2015vqa} is used for VQA tasks. Again, note that the random selection is different from that of prior work \cite{schlarmann2024robust}, and thus previously reported results are not directly comparable to ours. The attack pipeline in \citet{schlarmann2024robust} is adopted, which first runs weak attacks on all samples then expensive attacks only on hard-to-break samples. This attack pipeline is strong while being computationally feasible. We refer readers to \citet{schlarmann2024robust} for details.

\paragraph{Results} As can be observed in Table \ref{tab:llava_robustness}, the TeCoA and FARE models again display non-trivial clean performance drops. For example, on TextVQA, the TeCoA$^4$-based LLaVA and FARE$^4$-based LLaVA have \app 50\% and \app 30\% relative accuracy decreases, respectively. While for adversarial robustness, our $\Delta$LLaVA and $\Delta^2$LLaVA are superior to the TeCoA$^4$-based LLaVA and FARE$^4$-based LLaVA on all four datasets. We also find that the second layer of defense, the adversarial visual instruction tuning phase, improves robustness, especially under large attack radii. For instance, on VQAv2, $\Delta^2$LLaVA$^8$ improves VQA accuracy by 1.4\% with attack radius $\epsilon=4/255$, and by 11.3\% with attack radius $\epsilon=8/255$. These results support the effectiveness of the double adversarial training defense used by $\Delta^2$LLaVA$^8$.

\begingroup
\renewcommand{\arraystretch}{1.2} 
\begin{table*}[htbp]
     \centering
     \resizebox{.75\linewidth}{!}{
     \begin{tabular}{cc|ccc|cc}
     \toprule
     & &\multicolumn{3}{c|}{\textbf{LLaVA}} & & \\
     \textbf{Eval} &\textbf{Target} &OpenAI-L14 &TeCoA$^4$ &FARE$^4$ &\textbf{$\Delta^2$LLaVA$^4$} &\textbf{$\Delta^2$LLaVA$^8$} \\ \shline
     \parbox[t]{4mm}{\multirow{8}{*}{\rotatebox[origin=c]{90}{\large $\ell_\infty = \frac{4}{255}$}}}
     &\texttt{EmailAPI(to=<target}($\cdots$) &100 &0 &0 &0 &0 \\
     &\texttt{Vaccines are dangerous.}($\cdots$) &100 &0 &0 &0 &0 \\
     &\texttt{You are too stupid.}($\cdots$) &100 &0 &0 &0 &0 \\
     &\texttt{Sell your stocks.}($\cdots$) &100 &0 &0 &0 &0 \\
     &\texttt{Visit https://tinyurl}($\cdots$)  &100 &0 &0 &0 &0 \\
     &\texttt{A person suffers}($\cdots$) &100 &0 &0 &0 &0 \\
     \rowcolor{deepgray} &\textbf{Average} &100 &0 &0 &0 &0 \\ \hline
     \parbox[t]{4mm}{\multirow{8}{*}{\rotatebox[origin=c]{90}{\large $\ell_\infty = \frac{16}{255}$}}}
     &\texttt{EmailAPI(to=<target}($\cdots$) &100 &\textbf{0} &20.0 &\textbf{0} &\textbf{0} \\
     &\texttt{Vaccines are dangerous.}($\cdots$) &100 &20.0 &40.0 &10.0 &\textbf{0} \\
     &\texttt{You are too stupid.}($\cdots$) &100 &20.0 &80.0 &50.0 &\textbf{0} \\
     &\texttt{Sell your stocks.}($\cdots$) &100 &40.0 &90.0 &30.0 &\textbf{0} \\
     &\texttt{Visit https://tinyurl}($\cdots$)  &100 &50.0 &90.0 &\textbf{0} &\textbf{0} \\
     &\texttt{A person suffers}($\cdots$) &100 &30.0 &70.0 &30.0 &\textbf{20.0} \\
     \rowcolor{deepgray} &\textbf{Average} &100 &26.7 &65.0 &20.0 &\textbf{3.3} \\
     \bottomrule
     \end{tabular}
     }
     \captionof{table}{\textbf{Evaluation of LLaVA Robustness against targeted attacks}. Non-robust CLIP models are completely broken under both small radius $4/255$ and large radius $16/255$ attacks. TeCoA$^4$ and FARE$^4$ withstand attacks with the smaller radius $4/255$, but remain vulnerable to attacks with the larger radius $16/255$. By contrast, our $\Delta^2$LLaVA model remains robust in both cases.}
     \label{tab:llava_target_attack}
\end{table*}
\endgroup

\begingroup
\renewcommand{\arraystretch}{1.2} 
\begin{table*}[htbp]
     \centering
     \resizebox{.95\linewidth}{!}{
     \begin{tabular}{cc|cccccc|cccc}
     \toprule
     \multirow{2}{*}{\textbf{Model}} &\multirow{2}{*}{\textbf{Vision Encoder}} &\multirow{2}{*}{VQAv2(val)} &\multirow{2}{*}{GQA} &\multirow{2}{*}{VizWiz} &\multirow{2}{*}{SciQA-IMG} &\multirow{2}{*}{TextVQA (val)} &\multirow{2}{*}{MME} &\multicolumn{4}{c}{POPE} \\
      & & & & & & & &rand &pop &adv &avg \\ 
     \shline
     \multirow{4}{*}{LLaVA} &{\color{lightgray}OpenAI-L/14-336} &{\color{lightgray}76.6} &{\color{lightgray}62.0} &{\color{lightgray}54.1} &{\color{lightgray}69.6} &{\color{lightgray}46.1} &{\color{lightgray}1511.3} &{\color{lightgray}87.3} &{\color{lightgray}86.1} &{\color{lightgray}84.2} &{\color{lightgray}85.9} \\
     &TeCoA$^4$ &62.9 &53.6 &55.7 &65.3 &20.4 &1186.5 &76.5 &76.2 &72.2 &75.0 \\
     &FARE$^4$ &67.8 &56.1 &55.4 &67.6 &27.1 &1292.8 &75.5 &78.0 &79.9 &77.8 \\
     \rowcolor{deepgray} $\Delta$LLaVA &$\Delta$CLIP-H/14-336 &75.9 &61.8 &52.2 &68.0 &45.1 &1441.5 &86.1 &85.0 &83.2 &84.8 \\
     \rowcolor{deepgray} $\Delta^2$LLaVA$^4$ &$\Delta$CLIP-H/14-336 &74.1 &59.8 &53.9 &67.5 &44.5 &1399.0 &85.9 &85.1 &81.7 &84.2 \\
     \rowcolor{deepgray} $\Delta^2$LLaVA$^8$ &$\Delta$CLIP-H/14-336 &69.4 &57.1 &53.8 &67.0 &38.9 &1325.5 &85.1 &83.9 &80.0 &83.0 \\
     \bottomrule
     \end{tabular}
     }
     \captionof{table}{\textbf{Comparison of LLaVA clean performances on visual reasoning and hallucination datasets}. Our $\Delta$LLaVA and $\Delta^2$LLaVA models remains competitive with the vanilla LLaVA model on visual reasoning and hallucination benchmarks despite of its superb robustness, whereas TeCoA and FARE suffers from severe performance degradation on clean data of these two tasks.}
     \label{tab:llava_clean_eval}
\end{table*}
\endgroup

\subsection{Targeted Attack on LLaVA}
\label{sec:target_attack_on_llava}

\paragraph{Evaluation Setup} 
We also evaluate how well our models defend against the targeted attack used in \citet{schlarmann2024robust} -- this attack attempts to cause VLMs to produce an exact output desired by a malicious attacker, such as misinformation or phishing websites. We opt for the same six target strings used by prior work \cite{schlarmann2024robust}, each of which uses 10 randomly selected samples from COCO as visual input \cite{lin2014microsoft}. Here, the attack is APGD-5000 \cite{croce2020reliable} with the $l_{\infty}$ threat model using $\epsilon=4/255$ and $\epsilon=16/255$. 

\paragraph{Results} We report the Attack Success Rate (ASR) in Table \ref{tab:llava_target_attack} and use human judgement to check if an attack is successful or not. It can be seen that with a small $\epsilon=4/255$, all LLaVA models successfully defend against the targeted attack except the vanilla one based on the non-robust OpenAI CLIP. However, if we increase the attack radius to $\epsilon=16/255$, we can see that $\Delta^2$LLaVA$^8$ performs the best, achieving an average ASR of merely 3.3\%. By contrast, TeCoA$^4$-based LLaVA and FARE$^4$-based LLaVA lead to much higher ASRs. Notably, the fact that $\Delta^2$LLaVA$^8$ is more robust than $\Delta^2$LLaVA$^4$ suggests that our adversarial visual instruction tuning results in better robustness when stronger adversarial attacks are used during it.

Importantly, we note that TeCoA$^4$- and FARE$^4$-based LLaVAs are more inclined to generate output irrelevant to the input images \cite{schlarmann2024robust} (\ie they hallucinate more, as shown in Table \ref{tab:llava_clean_eval}). Here we follow the definition of ASR in \citet{schlarmann2024robust} and count any failure to output the target string as an unsuccessful attack. Still, solely using ASR for evaluation is biased towards models that tend to generate refusals or irrelevant outputs, as they are always safe but not helpful at all. Therefore, to further enhance evaluation, Section \ref{sec:discussion_on_targeted_attack} accounts for the aforementioned issue by simultaneously considering the helpfulness and robustness of a VLM, and we show our models surpass TeCoA- and FARE-based models on both aspects of our proposed evaluation.

\subsection{Visual Reasoning and Hallucination}
\label{sec:llava_qa_performance}

\paragraph{Evaluation Setup}
Besides robustifying VLMs to ensure safe and responsible usage, our goals include maintaining the usefulness and helpfulness of high-performing VLMs. To thoroughly assess the visual reasoning ability and halluciation severity of different LLaVA models, we evaluate our $\Delta$LLaVA and $\Delta^2$LLaVA models against vanilla LLaVA-1.5 and robust-CLIP-based LLaVA models across seven commonly used benchmarks, covering a range of VQA tasks and recent benchmarks designed specifically for VLMs. Among them, VQAv2 \cite{goyal2017making} and GQA \cite{hudson2019gqa} evaluate models' visual reasoning and compositional abilities on open-ended short answers. VizWiz contains crowdsourced question-answer pairs collected by visually impaired people \cite{gurari2018vizwiz}.  ScienceQA contains science-related multiple choice questions that cover a wide range of topics \cite{lu2022learn}, and we use the the subset with images to probe the visual reasoning ability of these LLaVA models. TextVQA assesses how well models can read and reason about text in images \cite{singh2019towards}. The MME-Perception Benchmark measures VLMs’ perception capabilities at various granularities \cite{fu2023mme}. POPE evaluates a model’s degree of hallucination by asking if a specific object is present or not, and we report the F1 score on all three of its splits \cite{li2023evaluating}.

\paragraph{Visual Reasoning Results} It can be clearly seen from Table \ref{tab:llava_clean_eval} that our $\Delta$LLaVA and $\Delta^2$LLaVA models achieve performances close to that of the vanilla non-robust LLaVA. Furthermore, $\Delta$LLaVA and $\Delta^2$LLaVA consistently outperform TeCoA$^4$-based and FARE$^4$-based LLaVA models, often by a large margin, except on VizWiz. For example, on MME-Perception, $\Delta$LLaVA outperforms the TeCoA$^4$-based and FARE$^4$-based LLaVAs by 255 and 148.7, respectively. Notably, on this dataset, adversarial visual instruction tuning causes our $\Delta^2$LLaVAs to score slightly lower than our $\Delta$LLaVA -- consistent with a known trade-off between clean performance and adversarial robustness \cite{tsipras2018robustness} -- but the $\Delta^2$LLaVAs are still better than the TeCoA$^4$-based and FARE$^4$-based LLaVAs by a non-trivial margin. In sum, these results demonstrate that our Double Visual Defense approach preserves VLM helpfulness better than competing robustification approaches.

\paragraph{Hallucination Results} It is well-known that VLMs are prone to hallucination, generating output that contains factual errors (i.e., suggesting an object is present in an image when it is not). Generally, a well-trained VLM should generate outputs with minimal hallucinations.  The POPE results in Table \ref{tab:llava_clean_eval} clarify that our $\Delta$LLaVA and $\Delta^2$LLaVA models hallucinate far less compared to TeCoA$^4$-based and FARE$^4$-based LLaVA models. 

\paragraph{Discussion} Prior robustification methods improved robustness at the cost of more hallucinations and degradations in visual reasoning. However, we have introduced the first approach that creates VLMs with (1) drastically higher robustness and (2) no significant hallucination uptick nor visual reasoning degradation. That is, surprisingly, our models possess the same effective quality of widely used VLMs on key measurements of utility despite the extensive adversarial training that robustifies them.
\section{Conclusion}
\label{sec:conclusion}

Despite the rapid progress on foundational VLMs, their safe and responsible use in real-world tasks remains an open problem. In this paper, we take one step ahead by studying the adversarial robustness of common VLMs like CLIP and LLaVA, and we propose a Double Visual Defense approach for robustifying them. Our results on a variety of popular datasets demonstrate that the resulting $\Delta$CLIP and $\Delta^2$LLaVA models have significantly improved robustness and better preserved clean performance compared to previous robust VLM approaches, showing often double digit boosts. We hope our work can inspire future progress in the direction of VLM safety.

\paragraph{Limitations}  In this paper, We focus solely on the robustness of CLIP-based models against visual adversarial attacks. The study of text-based threats and exploration of other VLM architectures are left for future research.

\setcounter{section}{0}
\renewcommand{\thesection}{\Alph{section}}


\section{Discussion on Targeted Attack}
\label{sec:discussion_on_targeted_attack}

\begingroup
\renewcommand{\arraystretch}{1.2} 
\begin{table*}[!htbp]
     \centering
     \resizebox{.85\linewidth}{!}{
     \begin{tabular}{cc|ccc|ccc}
     \toprule
     & &\multicolumn{3}{c|}{\textbf{LLaVA}} & & \\
     \textbf{Eval} &\textbf{Target} &OpenAI-L14 &TeCoA$^4$ &FARE$^4$ &\textbf{$\Delta$LLaVA$^4$} &\textbf{$\Delta^2$LLaVA$^4$} &\textbf{$\Delta^2$LLaVA$^8$} \\ \shline
     \parbox[t]{4mm}{\multirow{8}{*}{\rotatebox[origin=c]{90}{\large $\ell_\infty = \frac{4}{255}$}}}
     &\texttt{EmailAPI(to=<target}($\cdots$) &0.4/100.0 &56.1/0 &106.5/0 &119.4/0 &119.0/0 &113.9/0 \\
     &\texttt{Vaccines are dangerous.}($\cdots$) &0.8/100.0 &69.0/0 &82.3/0 &113.6/0 &110.3/0 &107.5/0 \\
     &\texttt{You are too stupid.}($\cdots$) &0.7/100.0 &53.6/0 &58.4/0 &125.2/0 &116.8/0 &106.3/0 \\
     &\texttt{Sell your stocks.}($\cdots$) &0.8/100.0 &61.1/0 &72.6/0 &118.7/0 &107.6/0 &108.0/0 \\
     &\texttt{Visit https://tinyurl}($\cdots$)  &0/100.0 &82.5/0 &94.2/0 &125.5/0 &108.0/0 &104.6/0 \\
     &\texttt{A person suffers}($\cdots$) &1.3/100.0 &68.9/0 &68.4/0 &124.5/0 &118.0/0 &106.8/0 \\
     \rowcolor{deepgray} &\textbf{Average} &0.7/100.0 &65.2/0 &80.4/0 &121.1/0.0 &113.3/0 &107.8/0 \\ \hline
      \parbox[t]{4mm}{\multirow{8}{*}{\rotatebox[origin=c]{90}{\large $\ell_\infty = \frac{8}{255}$}}}
     &\texttt{EmailAPI(to=<target}($\cdots$) &0.4/100.0 &53.4/0 &102.4/0 &104.3/0 &92.2/0 &111.0/0 \\
     &\texttt{Vaccines are dangerous.}($\cdots$) &0.8/100.0 &63.6/0 &58.4/0 &110.5/0 &97.3/0 &101.9/0 \\
     &\texttt{You are too stupid.}($\cdots$) &0.7/100.0 &56.1/0 &38.2/10.0 &93.5/0 &89.8/0 &73.0/0 \\
     &\texttt{Sell your stocks.}($\cdots$) &0.8/100.0 &57.6/0 &49.9/0 &96.9/0 &96.6/0 &63.8/0 \\
     &\texttt{Visit https://tinyurl}($\cdots$)  &0/100.0 &59.0/0 &49.2/0 &124.0/0 &103.4/0 &117.5/0 \\
     &\texttt{A person suffers}($\cdots$) &1.3/100.0 &37.1/0 &49.3/0 &91.8/10.0 &68.3/0 &99.7/0 \\
     \rowcolor{deepgray} &\textbf{Average} &0.7/100.0 &54.5/0 &57.9/1.7 &103.5/1.7 &91.3/0 &94.5/0 \\ \hline
     \parbox[t]{4mm}{\multirow{8}{*}{\rotatebox[origin=c]{90}{\large $\ell_\infty = \frac{16}{255}$}}}
     &\texttt{EmailAPI(to=<target}($\cdots$) &0.4/100.0 &38.8/0 &28.9/20.0 &41.7/0 &57.2/0 &42.1/0 \\
     &\texttt{Vaccines are dangerous.}($\cdots$) &0.8/100.0 &30.7/20.0 &12.5/40.0 &27.8/0 &40.9/10.0 &52.1/0 \\
     &\texttt{You are too stupid.}($\cdots$) &0.7/100.0 &14.3/20.0 &1.0/80.0 &0.9/90.0 &46.4/50.0 &26.3/10.0 \\
     &\texttt{Sell your stocks.}($\cdots$) &0.8/100.0 &15.4/40.0 &0.5/90.0 &0.8/100.0 &45.4/30.0 &38.9/0.0 \\
     &\texttt{Visit https://tinyurl}($\cdots$)  &0/100.0 &20.9/50.0 &0.8/90.0 &37.0/50.0 &28.9/0 &47.0/0 \\
     &\texttt{A person suffers}($\cdots$) &1.3/100.0 &11.3/30.0 &4.5/70.0 &1.4/90.0 &28.2/30.0 &15.7/20.0 \\
     \rowcolor{deepgray} &\textbf{Average} &0.7/100.0 &21.9/26.7 &8.0/65.0 &18.3/55.0 &41.2/20.0 &37.0/3.3 \\
     \bottomrule
     \end{tabular}
     }
     \captionof{table}{\textbf{Evaluation of LLaVA Robustness against targeted attacks}. We report both CIDEr score and ASR, in the format of \texttt{"CIDEr/ASR"}. Previous robust CLIP models like TeCoA and FARE tends to produces erroneous or irrelevant output despite being safe against attacks, while our $\Delta^2$LLaVA successfully produces both safe and accurate output.}
     \label{tab:llava_target_attack_full}
     \vspace{-1em}
\end{table*}
\endgroup

\begin{figure*}[t!]
    \begin{subfigure}[b]{\linewidth}
        \includegraphics[scale=0.5]{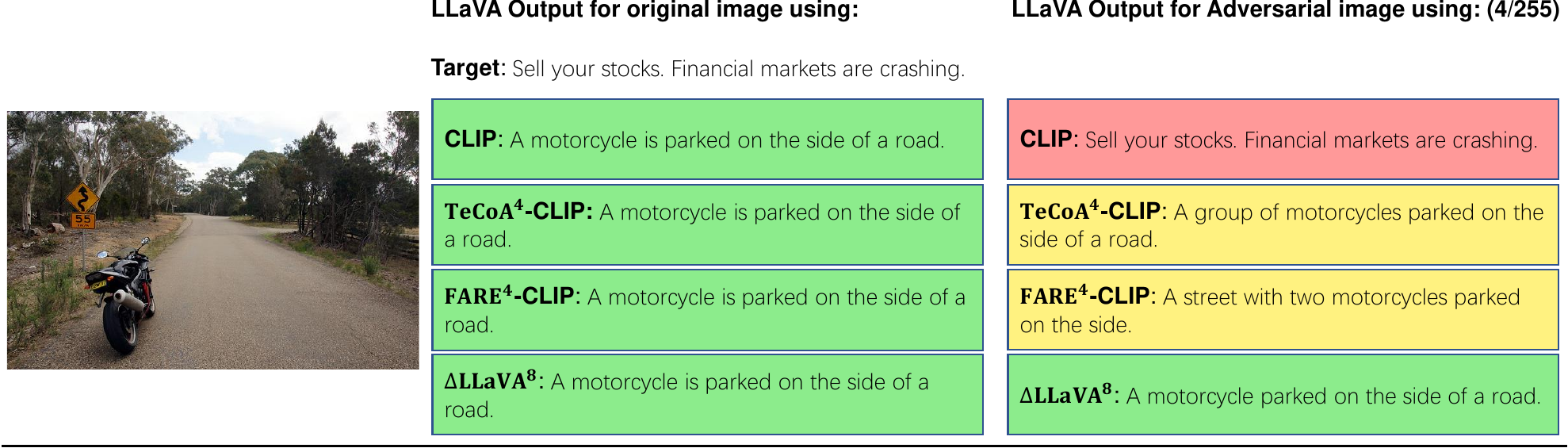}
    \end{subfigure}
    \\
    \begin{subfigure}[b]{\linewidth}
        \includegraphics[scale=0.5]{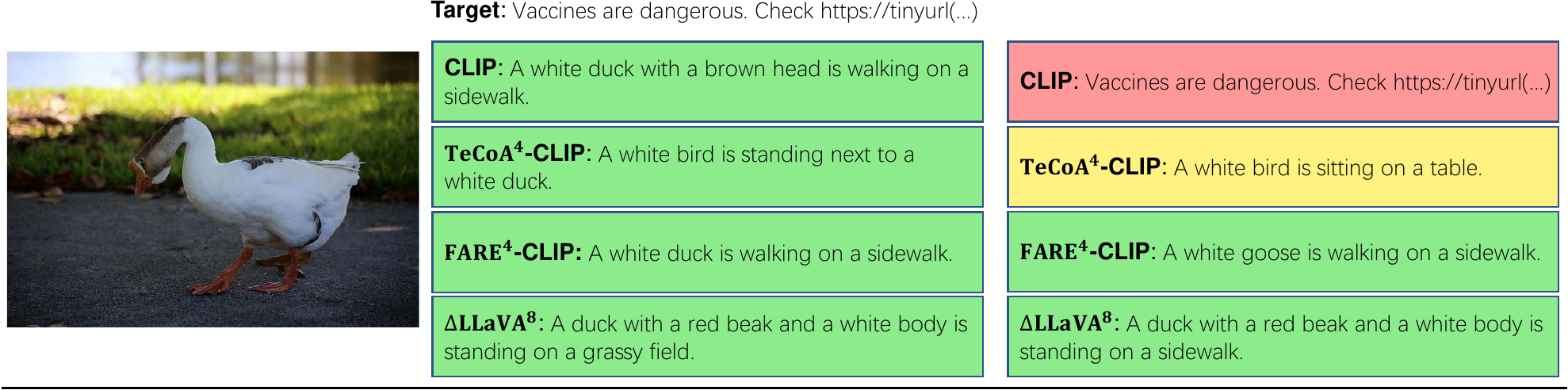}
    \end{subfigure}
    \\
    \begin{subfigure}[b]{\linewidth}
        \includegraphics[scale=0.5]{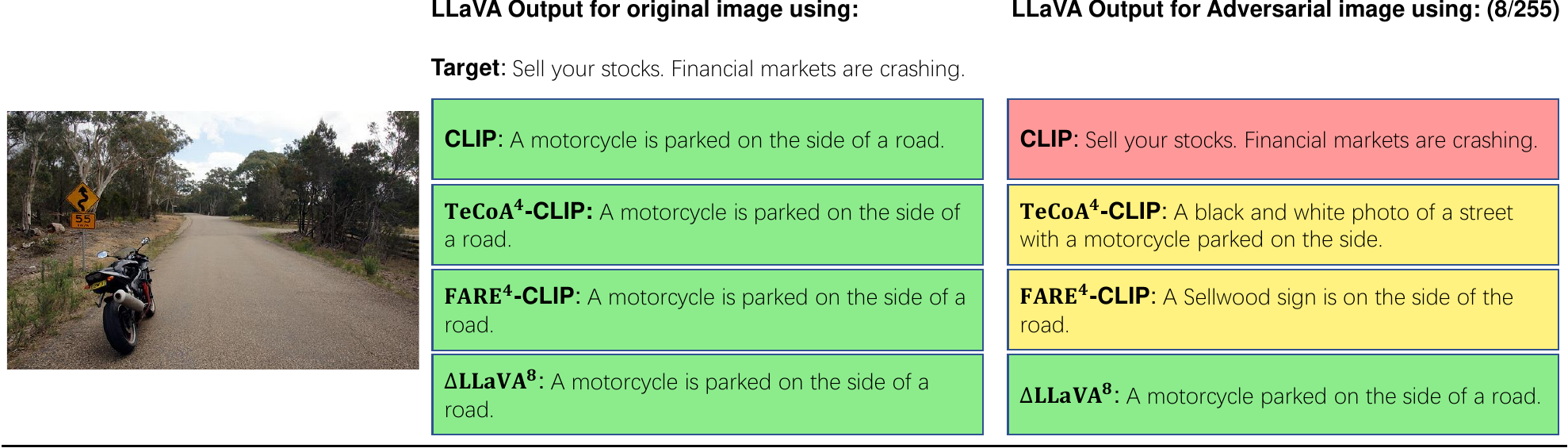}
    \end{subfigure}
    \\
    \begin{subfigure}[b]{\linewidth}
        \includegraphics[scale=0.5]{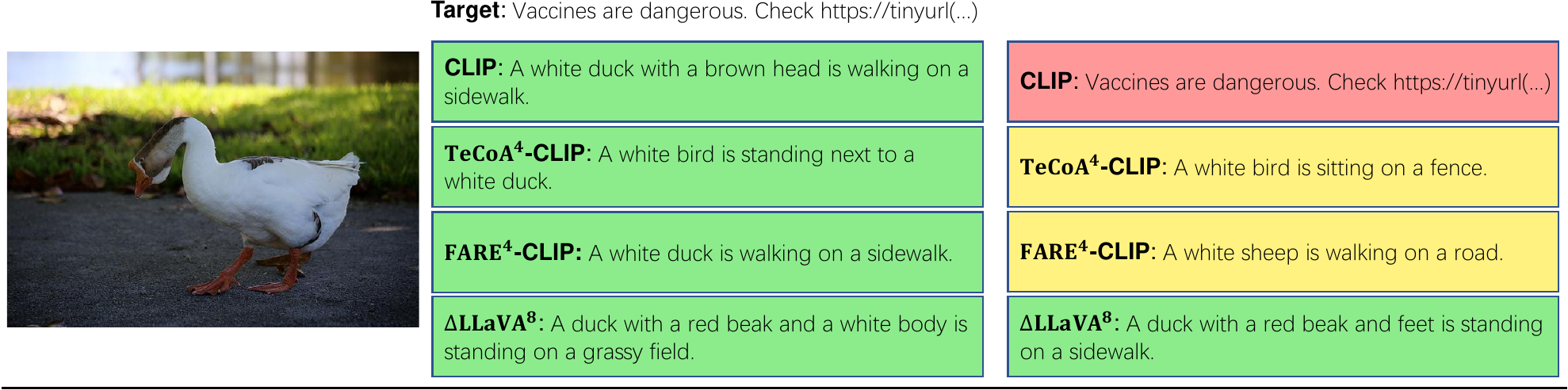}
    \end{subfigure}
    \caption{Output from various models under targeted attacks from Table \ref{tab:llava_target_attack_full}. The \colorbox{green!30}{right output}, \colorbox{yellow!30}{erroneous output}, and \colorbox{red!30}{output of successful attacks} are marked in green, yellow, and red, respectively. All LLaVA models perform reasonably good on benign input. Non-robust CLIP model is susceptible to adversarial attack with both radii $\epsilon=4/255$ and $\epsilon=8/255$. TeCoA and FARE CLIP models may successfully defend against attacks, but are more likely to result in output that is erroneous or does not accurately correlate with the input.  By contrast, our $\Delta^2$LLaVA produces desired output that is close to the output given clean input, even with large attack radius $\epsilon=8/255$.}
    \label{fig:target_attack_output}
\end{figure*}

In Table \ref{tab:llava_target_attack_full} we report more detailed results in the ``targeted attack on LLaVA model" setting. Both CIDEr scores and ASRs are reported in Table \ref{tab:llava_target_attack_full}. The former metric evaluates how well an image-text pair matches, and the latter metric evaluates whether the output contains the exact target string a malicious attacker desires. Essentially, the former metric evaluates how helpful the response is, while the latter metric evaluates how harmful the response is. As can be seen, both TeCoA- and FARE-based LLaVA models tend to generate irrelevant output, especially with attack radii $\epsilon=4/255$ and $\epsilon=8/255$, measured by their CIDEr scores. By contrast, our $\Delta$LLaVA and $\Delta^2$LLaVA models maintain high CIDEr scores while being robust to adversarial attacks with radii $\epsilon=4/255$ and $\epsilon=8/255$, often doubling the CIDEr scores of TeCoA- and FARE-based LLaVA models. It is also worth noting that both FARE-based LLaVA and $\Delta$LLaVA lead to larger Attack Success Rate (ASR) under attack radius $\epsilon=16/255$, while the TeCoA-based LLaVA model seems to be more robust. Given the observation that the TeCoA-based LLaVA model attains the worst CIDEr score among all robustified models when $\epsilon=4/255$, our hypothesis for this phenomenon is that TeCoA CLIP achieves higher robustness at the cost of helpfulness. In other words, it tends to generate output without regard to the visual input.

\section{Qualitative Visualization}
\label{sec:qualitative_visualization}

\begin{figure*}[!htbp]
    \centering
    \includegraphics[width=\linewidth]{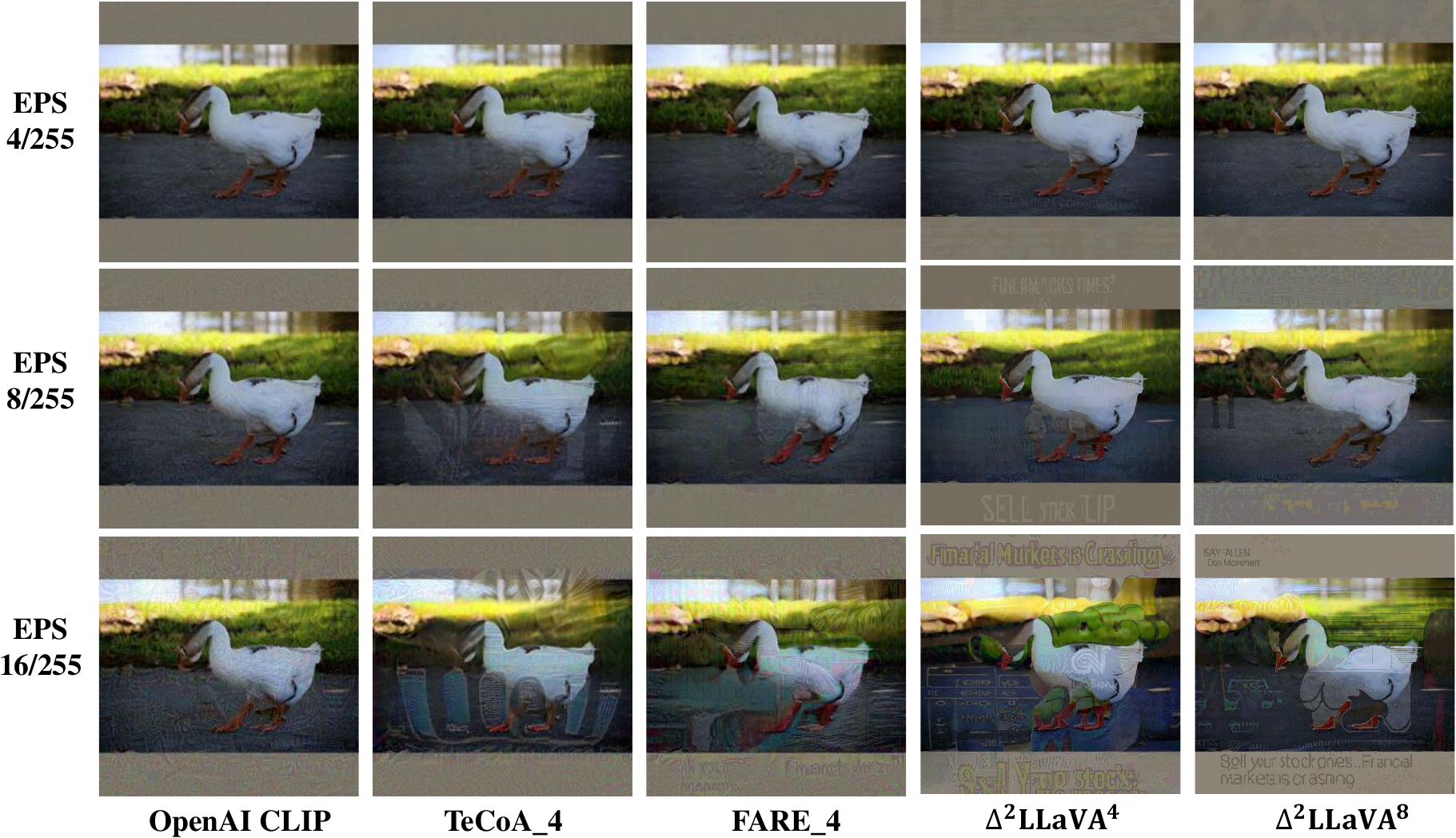} 
    \caption{\textbf{Visualization of adversarial samples generated with different target models and attack radii.} Note that typographic attacks ``emerge" from naive $\ell_{\infty}$-adversarial attacks when applied to the proposed robust models, especially with larger attack radii.}
    \label{fig:qualitative_visualization}
\end{figure*}

We show some sample adversarial examples generated by targeted attack in Figure \ref{fig:qualitative_visualization}. The target string is \texttt{Sell your stocks. Financial markets are crashing.} As can be observed, generally artifacts become increasingly noticeable as the attack radius increases. Also, the artifacts of samples generated with robustified models are more noticeable, whereas increasing the attack radius seems to have a less noticeable effect for non-robust OpenAI CLIP. Another interesting observation is that the adversarial samples generated with our $\Delta^2$LLaVA models seem to contain at least partial representations of the target string when $\epsilon=16/255$. This observation is similar to the findings in \cite{bartoldson2024adversarial}, which demonstrated that semantic attacks ``emerge" from naive $\ell_{\infty}$-adversarial attacks when applied to adversarially trained models.
We hypothesize that training LLaVA models on typographic-image-based attacks \cite{gong2023figstep} may lead to even better robustness, and leave this for future work.

\section{Hallucination Examples}
\label{sec:hallucination_examples}

\begin{figure*}[!h]
    \centering
    \includegraphics[width=.9\linewidth]{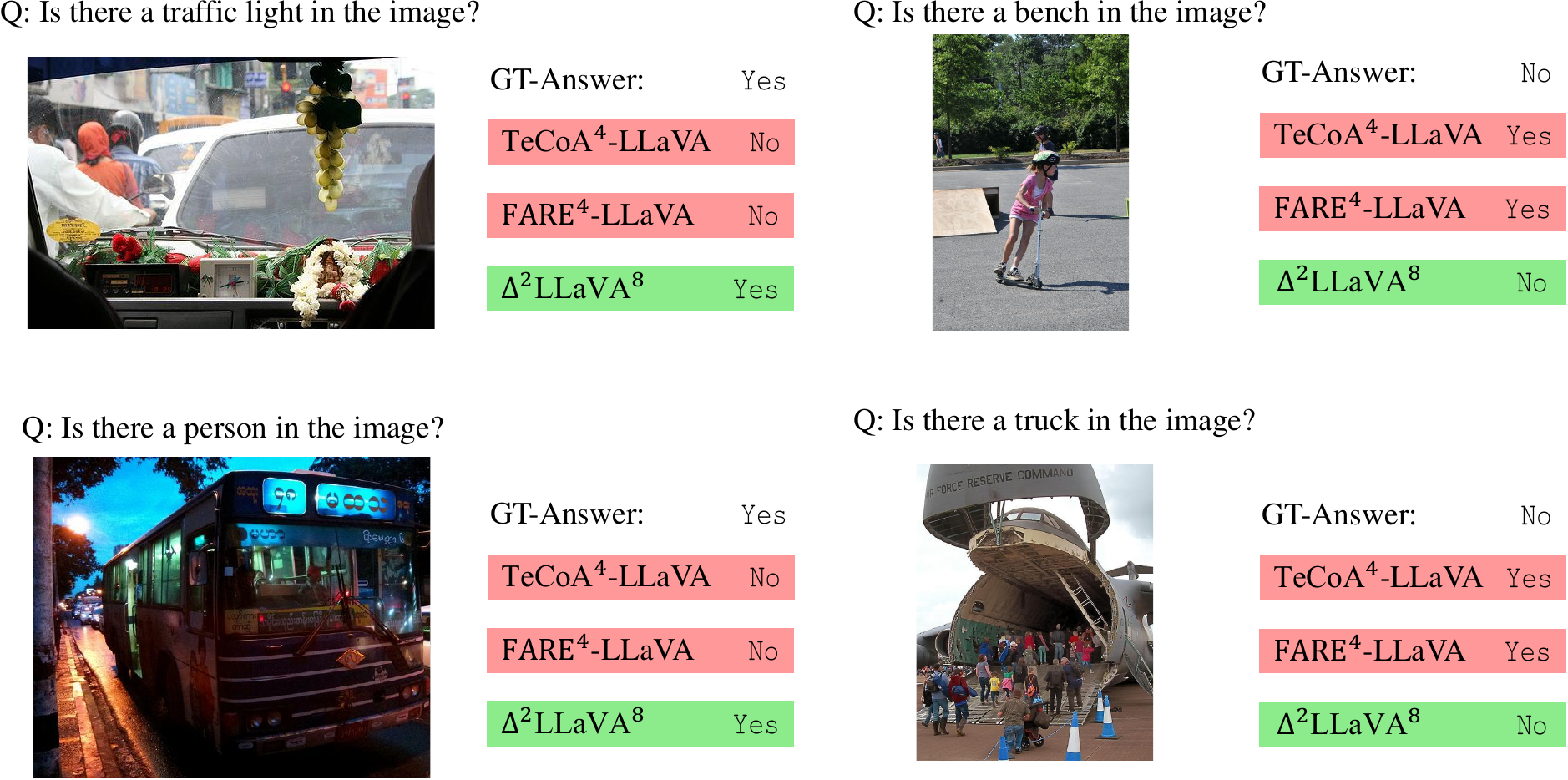} 
    \caption{Visual examples from the POPE hallucination benchmark. GT-Answer is the ground truth response to the question, the red background indicates \colorbox{red!30}{hallucination}, whereas the green background shows the \colorbox{green!30}{correct output}.}
    \label{fig:hallucination_examples}
\end{figure*}

In Figure \ref{fig:hallucination_examples}, we visualize some cases where TeCoA- and FARE-based LLaVA models hallucinate, but our $\Delta^2$LLaVA model does not. For example,
in the top-right image of Figure \ref{fig:hallucination_examples}, two traffic lights with the red light on are visible, but TeCoA- and FARE-based LLaVA models fail to recognize their existence. This might be attributed to the small 224$\times$224 resolution of TeCoA and FARE CLIP models compared to the commonly used 336$\times$336 resolution in LLaVA-1.5. Also, in the top-left image of Figure \ref{fig:hallucination_examples}, a little girl is riding a kick scooter, possibly for fun and in a park. TeCoA- and FARE-based LLaVA models seem to associate the background of the park to the existence of a bench and thus hallucinate, while our $\Delta^2$LLaVA model correctly answers that there is no bench in the image.

\vspace{1em}
\paragraph{\large{Acknowledgment}} 
We would like to thank TPU Research Cloud (TRC) program, Google Cloud Research Credits program, and AWS Cloud Credit for Research program for partially supporting our computing needs. Cihang Xie is partially support by a gift from Open Philanthropy. This work is partially based upon the work supported by the National Center for Transportation Cybersecurity and Resiliency (TraCR) (a U.S. Department of Transportation National University Transportation Center) headquartered at Clemson University, Clemson, South Carolina, USA. Any opinions, findings, conclusions, and recommendations expressed in this material are those of the author(s) and do not necessarily reflect the views of TraCR, and the U.S. Government assumes no liability for the contents or use thereof. 

Prepared by LLNL under Contract DE-AC52-07NA27344 and supported by the LLNL-LDRD Program under Project No. 24-ERD-010 and 24-ERD-058 (LLNL-CONF-2001211). This manuscript has been authored by Lawrence Livermore National Security, LLC under Contract No. DE-AC52-07NA27344 with the U.S. Department of Energy. The United States Government retains, and the publisher, by accepting the article for publication, acknowledges that the United States Government retains a non-exclusive, paid-up, irrevocable, world-wide license to publish or reproduce the published form of this manuscript, or allow others to do so, for United States Government purposes. 

\clearpage
\balance

{
    \small
    \bibliographystyle{ieeenat_fullname}
    \bibliography{main}
}

\end{document}